\documentclass{article}

\usepackage[preprint]{neurips_2026}

\usepackage[utf8]{inputenc}
\usepackage[T1]{fontenc}
\usepackage{times}
\usepackage{microtype}
\usepackage{graphicx}
\usepackage{float}
\usepackage{placeins}
\usepackage{booktabs}
\usepackage{nicefrac}
\usepackage{makecell}
\usepackage{multirow}
\usepackage{array}
\usepackage{siunitx}
\usepackage[table,xcdraw,dvipsnames]{xcolor}
\usepackage{caption}
\usepackage{subcaption}
\usepackage{wrapfig}
\usepackage{enumitem}
\usepackage[most]{tcolorbox}
\usepackage{amsmath,amssymb,amsfonts,mathtools}
\usepackage{xspace}
\usepackage{pifont}
\usepackage{url}
\usepackage{threeparttable}
\usepackage{tabularx}
\usepackage{arydshln}
\usepackage{hyperref}
\usepackage{cleveref}

\newcommand{\llmname}[1]{{\texttt{#1}}\xspace}

\definecolor{aropdblue}{HTML}{2563EB}
\definecolor{aropdanchor}{HTML}{2E7D57}
\definecolor{aropdfull}{HTML}{D97706}
\definecolor{aropdresidual}{HTML}{7C3AED}
\definecolor{aropdstudent}{HTML}{64748B}
\definecolor{aropdrisk}{HTML}{DC2626}
\definecolor{aropdlightblue}{HTML}{CFE2FF}
\definecolor{aropdlightsecond}{HTML}{E8F1FF}
\definecolor{aropdlightgreen}{HTML}{E7F5EC}
\definecolor{aropdlightorange}{HTML}{FFF2D8}
\definecolor{aropdtablebase}{HTML}{F8FAFC}
\definecolor{aropdtablesft}{HTML}{F1F5F9}
\definecolor{aropdtablefull}{HTML}{FFF4E6}
\definecolor{aropdtableours}{HTML}{E8F7EE}
\definecolor{aropdtableavg}{HTML}{EAF2FF}

\newcommand{\hlfirst}[1]{\colorbox{aropdlightblue}{#1}}

\newcolumntype{L}[1]{>{\raggedright\arraybackslash}p{#1}}
\newcolumntype{C}[1]{>{\centering\arraybackslash}p{#1}}

\newcommand{\student}{\pi_\theta}
\newcommand{\teacher}{\pi_{\bar{\theta}}}
\newcommand{\qpart}{q_{\mathrm{part}}}
\newcommand{\qfull}{q_{\mathrm{full}}}
\newcommand{\qlambda}{q_{\lambda}}

\newcommand{\kl}{D_{\mathrm{KL}}}

\newcommand{\anchorname}{\textcolor{aropdanchor}{\textsc{Anchor}}\xspace}
\newcommand{\fullname}{\textcolor{aropdfull}{\textsc{Full View}}\xspace}
\newcommand{\residualname}{\textcolor{aropdresidual}{\textsc{Residual}}\xspace}

\newcommand{\anchorterm}[1]{\textcolor{aropdanchor}{#1}}
\newcommand{\fullterm}[1]{\textcolor{aropdfull}{#1}}

\newtcolorbox{aropdquestionbox}{
  enhanced,
  colframe=black!70,
  colback=aropdlightblue!35,
  boxrule=0.8pt,
  arc=2mm,
  left=5pt,
  right=5pt,
  top=5pt,
  bottom=5pt
}

\newtcolorbox{aropdtakeawaybox}{
  enhanced,
  colframe=aropdblue!75!black,
  colback=aropdlightblue!25,
  boxrule=0.8pt,
  arc=2mm,
  left=5pt,
  right=5pt,
  top=5pt,
  bottom=5pt
}

\newtcolorbox{aropdvisualbox}{
  enhanced,
  colframe=aropdblue!45!black,
  colback=white,
  boxrule=0.7pt,
  arc=1.5mm,
  left=6pt,
  right=6pt,
  top=6pt,
  bottom=6pt
}

\newtcolorbox{aropdmethodbox}{
  enhanced,
  colframe=aropdanchor!70!black,
  colback=aropdlightgreen!42,
  boxrule=0.75pt,
  arc=1.5mm,
  left=6pt,
  right=6pt,
  top=5pt,
  bottom=5pt
}

\newtcolorbox{aropdempiricalbox}{
  enhanced,
  colframe=aropdanchor!75!black,
  colback=aropdlightgreen!55,
  boxrule=0.8pt,
  arc=2mm,
  left=5pt,
  right=5pt,
  top=5pt,
  bottom=5pt
}

\newtcolorbox{aropdpromptbox}{
  enhanced,
  breakable,
  colframe=aropdresidual!70!black,
  colback=aropdlightsecond!45,
  boxrule=0.75pt,
  arc=1.5mm,
  left=7pt,
  right=7pt,
  top=6pt,
  bottom=6pt
}

\newcommand{\cnum}[1]{\ding{\numexpr181 + #1\relax}}
\newcommand{\aropditem}[1]{\item[\textcolor{aropdblue}{\cnum{#1}}]}

\hypersetup{
    colorlinks=true,
    linkcolor=aropdrisk,
    citecolor=aropdblue,
    urlcolor=aropdresidual,
}

\newcommand{\AROPDhighlight}[1]{%
\begin{aropdempiricalbox}
\textbf{Empirical highlight.} #1
\end{aropdempiricalbox}
}

\title{Beyond Absolute Imitation: Anchored Residual Guidance for Privileged On-Policy Distillation}

\author{%
  Wenhao Zhang\\
  South China University of Technology\\
  \texttt{vanhowe@outlook.com}
}

\begin{document}

\maketitle

\begin{abstract}
On-policy distillation (OPD) has demonstrated strong empirical gains in enhancing complex reasoning in LLMs by aligning a student model with a teacher's predictive distribution over the student's own trajectories. An emerging variant, Privileged OPD, further strengthens this paradigm by employing a self-teacher model augmented with privileged information, such as oracle traces, to mitigate teacher-student capacity gaps while providing dense, answer-directed supervision. However, current methods treat privileged information as a monolithic imitation target, failing to disentangle locally reachable reasoning steps from future-conditioned oracle signals. Consequently, the student is encouraged to match a hindsight-biased distribution that often falls outside its local predictive support. This reachability mismatch incentivizes the student model to skip valid intermediate reasoning in favor of locally unsupported shortcuts. To resolve this, we introduce \textbf{Anchored Residual On-Policy Distillation (AR-OPD)}, a dual-view framework that disentangles privileged supervision. Rather than enforcing strict full-view imitation, AR-OPD establishes a locally compatible anchor using a partially privileged teacher, isolating and injecting oracle foresight as a controlled residual to provide destination-directed guidance. Across diverse reasoning tasks, \textbf{AR-OPD outperforms full privileged OPD by 2.3 points and SFT by 7.9 points}. Crucially, this anchored residual mechanism reduces hindsight leakage by \textbf{21.7\%} and mitigates late-stage drift, yielding up to a \textbf{7.2-point advantage} on challenging long-horizon trajectories exceeding 768 tokens.
\end{abstract}

\section{Introduction}
\label{sec:intro}
On-policy distillation (OPD) has rapidly emerged as an effective paradigm for large language model post-training, driving substantial gains in recent industry pipelines~\citep{yang2025qwen3,xiao2026mimo,zeng2026glm5}. By querying a teacher on states actively induced by the student, OPD addresses the off-policy exposure bias of supervised fine-tuning~\citep{brown2020language,wei2022chain,ouyang2022instructgpt,taori2023alpaca} and the reward sparsity of reinforcement learning~\citep{schulman2017ppo,guo2025deepseekr1,rlhf-workflow,verl,thinkingmachines2025opd,agarwal2024policy}. However, standard cross-model OPD often suffers from pattern mismatch: distilling a structurally incompatible teacher can degrade student performance~\citep{li2026rethinkingopd,fu2026revisitingopd,jang2026veto}. 

To circumvent this, \textit{Privileged OPD} (e.g., OPSD, SDFT~\citep{zhao2026opsd}) restricts the teacher and student to the same base model but augments the teacher with auxiliary context, such as oracle traces. By allowing the self-teacher to ``see more,'' privileged OPD improves teacher-student compatibility and provides dense supervision aligned with the student's native generation manifold. While Privileged OPD densifies the learning signal, it can also shift the teacher's distribution away from the student's causal predictive support.

This creates a different failure mode from ordinary teacher-student mismatch. The teacher is not merely stronger or weaker; it is conditioned on information that the student cannot access at the same prefix. Thus, the central question is not whether privileged information is useful, but how much of it remains locally learnable as a token-level target.

\begin{figure}[!t]
\centering
\vspace{-0.4em}
\includegraphics[width=0.98\linewidth]{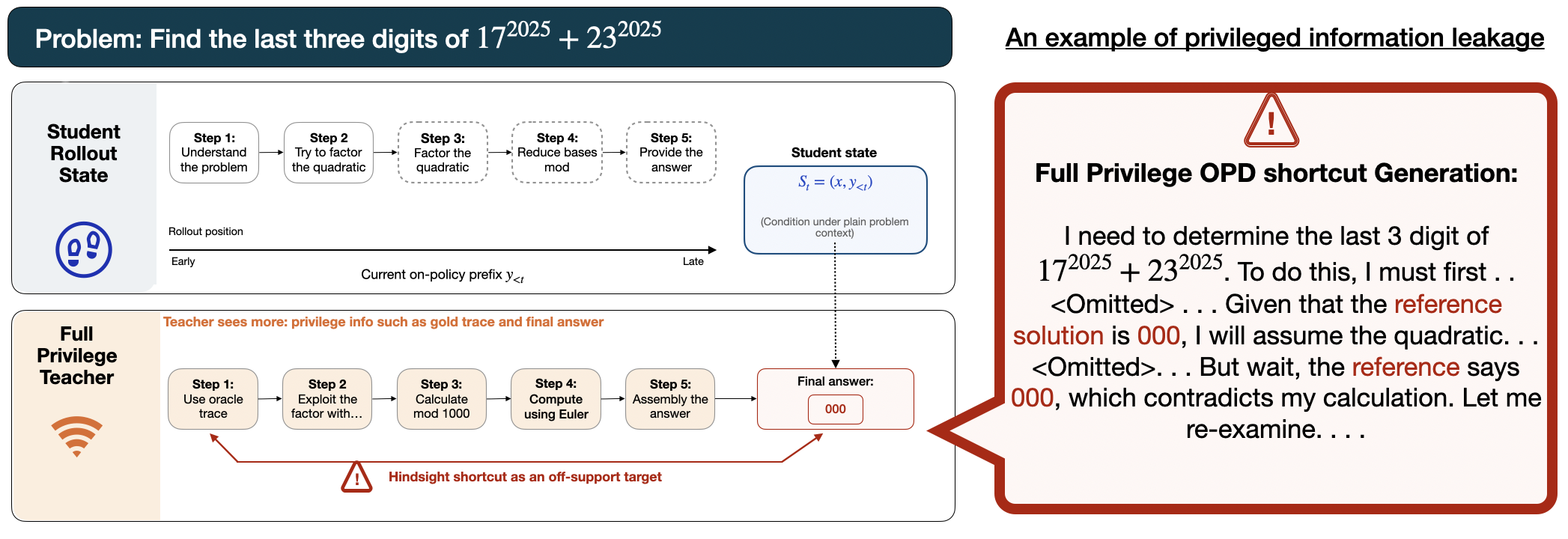}
\vspace{-0.6em}
\caption{\textbf{Privileged-information leakage as an off-support target.} A representative example illustrating privileged-information leakage in a privileged OPD-trained model, where the model appeals to an invisible reference solution during inference. Because the full privileged teacher conditions on oracle information unavailable to the student, it can assign probability mass to answer-conditioned states. Direct imitation therefore turns factually correct privileged information into a shortcut target.}
\label{fig:motivation}
\vspace{0.45em}
\end{figure}

Consequently, treating this fully privileged teacher as an absolute imitation target introduces a subtle but persistent optimization pathology. Because the full-view oracle conditions on the complete future context, its distribution implicitly incorporates hindsight-derived information and future-conditioned structure. It shifts probability mass toward tokens that, while factually correct, remain causally unsupported from the student's current prefix—such as anticipating a final answer branch before establishing the necessary intermediate rationale. We characterize these premature emissions as \textbf{privileged-information leakage} or \textit{shortcut} events (\Cref{fig:motivation} illustrates this target-side failure, where a privileged OPD-trained model appeals to an invisible ``reference solution'' during inference, a phenomenon later quantified in \Cref{fig:shortcut_accuracy}). In \Cref{sec:failure}, we show that this mismatch becomes most visible near the rollout tail, where full-view targets assign increasing mass outside the student's local support. We further show that full-view reliability itself degrades with privileged-context length, making monolithic full-view imitation brittle precisely in long-horizon reasoning.

To overcome this absolute imitation trap, we propose \textbf{Anchored Residual On-Policy Distillation (AR-OPD)}, an objective designed to extract privileged guidance while preserving target reachability. Rather than forcing the assimilation of a monolithic full teacher, AR-OPD disentangles privileged supervision into two controllable components: a locally compatible \anchorname and a destination-directed \residualname. The \anchorname uses a partially privileged teacher to provide a causally reachable target that isolates the student from hindsight-biased answers, while the \residualname injects the full-view teacher's marginal foresight as a \(\lambda\)-scaled log-probability update over the anchor. This shifts imitation from an unsupported hindsight coordinate to a controlled, future-directed update.

Our contributions are summarized as follows:
\vspace{-0.4em}
\begin{itemize}[leftmargin=2em,itemsep=-0.1em]
\aropditem{1} \textbf{Controlling the impact of privileged information.} We show that monolithic use of oracle-augmented teachers can induce optimization pathologies, and introduce a framework that extracts useful destination-directed foresight while controlling hindsight-biased manifold drift.

\aropditem{2} \textbf{Anchored Residual On-Policy Distillation (AR-OPD).} We introduce a dual-view distillation method that operationalizes this control through a causally reachable partial-oracle \anchorname and a \(\lambda\)-scaled log-space \residualname, transferring privileged guidance without sacrificing local sequence compatibility.

\aropditem{3} \textbf{Mechanistic evidence and cross-domain gains.} AR-OPD suppresses privileged-information leakage, reducing shortcut events by \textbf{21.7\%}. It also achieves the highest average score across diverse reasoning benchmarks, improving over the base model by \textbf{12.1 points} and full privileged OPD by \textbf{2.3 points}.

\end{itemize}

\clearpage
\begin{figure*}[!t]
\centering
\includegraphics[width=0.96\linewidth]{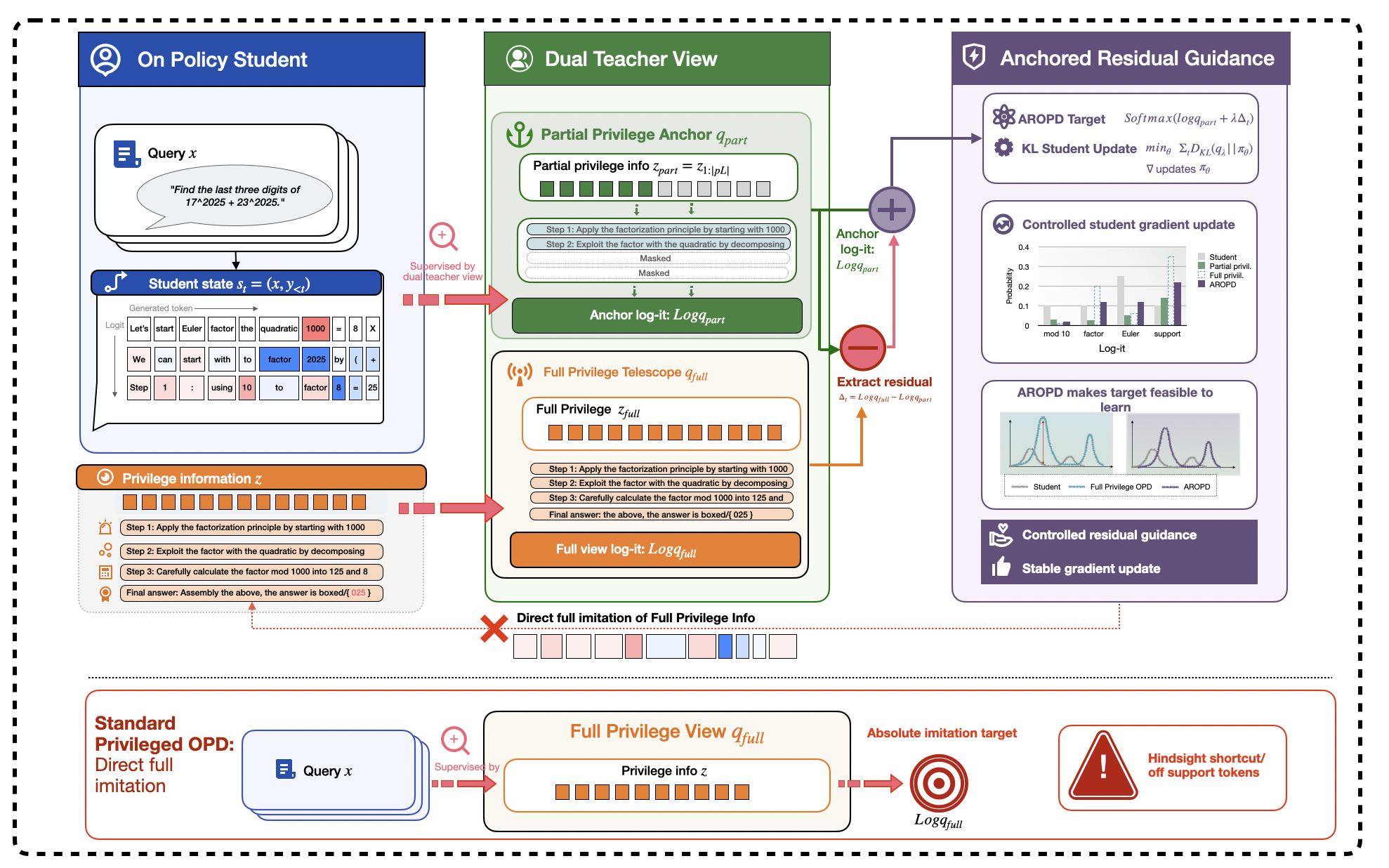}
\caption{\textbf{AR-OPD constructs a dual-view anchored residual target.} The \anchorname supplies the reachable reference, while the \residualname transfers controlled full-view guidance before distillation to the student.}
\label{fig:architecture}
\end{figure*}

\section{Related Work}
\label{sec:related}

\paragraph{Knowledge distillation and on-policy distillation.}
Knowledge distillation trains a student to match a teacher distribution or teacher-generated outputs~\citep{hinton2015distilling,seq-kd,distilbert,gu2023minillm}. For autoregressive generation, off-policy distillation and SFT can suffer from exposure mismatch because training targets are not conditioned on student-generated prefixes~\citep{ross2011dagger,lamb2016professor}. OPD addresses this by querying teachers on student-visited states and has become a strong post-training recipe for reasoning models~\citep{thinkingmachines2025opd,agarwal2024policy,fu2026revisitingopd,li2026rethinkingopd}. Recent variants study objective stability, entropy control, and teacher-student compatibility, showing that dense teacher targets are useful but can also become harmful when teacher distributions are locally incompatible with the student~\citep{jang2026veto,jin2026eaopd,yang2026gopd}.

\paragraph{Privileged OPD.}
Privileged OPD conditions the teacher on auxiliary information such as demonstrations, reference traces, verifier feedback, or oracle answers~\citep{lightman2023let,shenfeld2026sdft,zhao2026opsd,penaloza2026privileged,deepcritic}. Existing variants differ mainly in how they stabilize this privileged teacher. SDFT uses a delayed or EMA teacher to reduce training instability~\citep{shenfeld2026sdft}, whereas OPSD keeps teacher and student synchronized and adds full-vocabulary JSD with per-token clipping~\citep{zhao2026opsd}. Despite these differences, they still treat the privileged distribution as an absolute imitation target, leaving target reachability and hindsight-induced support mismatch largely unaddressed.

\section{Preliminaries}
\label{sec:prior}

\subsection{On-Policy Distillation}

In autoregressive OPD, prompts \(x \sim \mathcal{D}\) are paired with student rollouts \(y=(y_1,\ldots,y_T) \sim \student(\cdot \mid x)\). The state at step \(t\) is \(s_t=(x,y_{<t})\), and the student policy factorizes as
\begin{equation}
\student(y \mid x)=\prod_{t=1}^{T}\student(y_t \mid s_t).
\label{eq:policy_factorization}
\end{equation}
Given a teacher distribution \(q(\cdot \mid s_t)\), the target of OPD is to minimize token-level KL divergence on student-visited states:
\begin{equation}
\mathcal{L}_{\mathrm{OPD}}(\theta)
=
\mathbb{E}_{x\sim\mathcal{D},\,y\sim\student}
\left[
\sum_{t=1}^{T}
\kl\big(q(\cdot\mid s_t)\parallel \student(\cdot\mid s_t)\big)
\right].
\label{eq:opd}
\end{equation}
Different OPD implementations can provide full logits, top-\(k\) logits, or hard next-token labels. We use the logit-level view because it makes target construction observable: changing the privileged context changes the entire teacher distribution, not only the sampled next token.

\subsection{Privileged OPD}

Privileged OPD conditions the teacher on auxiliary information \(z\), such as a solution trace, an oracle answer, or an expert demonstration. With a frozen or delayed teacher parameter copy \(\bar{\theta}\), the privileged teacher is
\begin{equation}
q_z(\cdot\mid s_t)=\teacher(\cdot\mid x,z,y_{<t}).
\label{eq:priv_teacher}
\end{equation}
The student is trained without \(z\) at inference time. Direct privileged imitation therefore uses the training objective
\begin{equation}
\mathcal{L}_{\mathrm{P\mbox{-}OPD}}(\theta)
=
\mathbb{E}_{x\sim\mathcal{D},\,y\sim\student}
\left[
\sum_{t=1}^{T}
\kl\big(q_z(\cdot\mid s_t)\parallel\student(\cdot\mid s_t)\big)
\right].
\label{eq:pi_opd}
\end{equation}

\section{Why Full-View Imitation Fails Local Alignment}
\label{sec:failure}

Standard Privileged OPD assumes that oracle-augmented teachers provide uniformly reliable targets. We challenge this assumption by diagnosing the inherent reachability mismatch and signal fragility in absolute full-view imitation.

\subsection{Target Unreachability: The Hindsight Support Gap}

\begin{figure}[!t]
\centering
\vspace{0.2em}
\includegraphics[width=\linewidth,trim=4 8 4 8,clip]{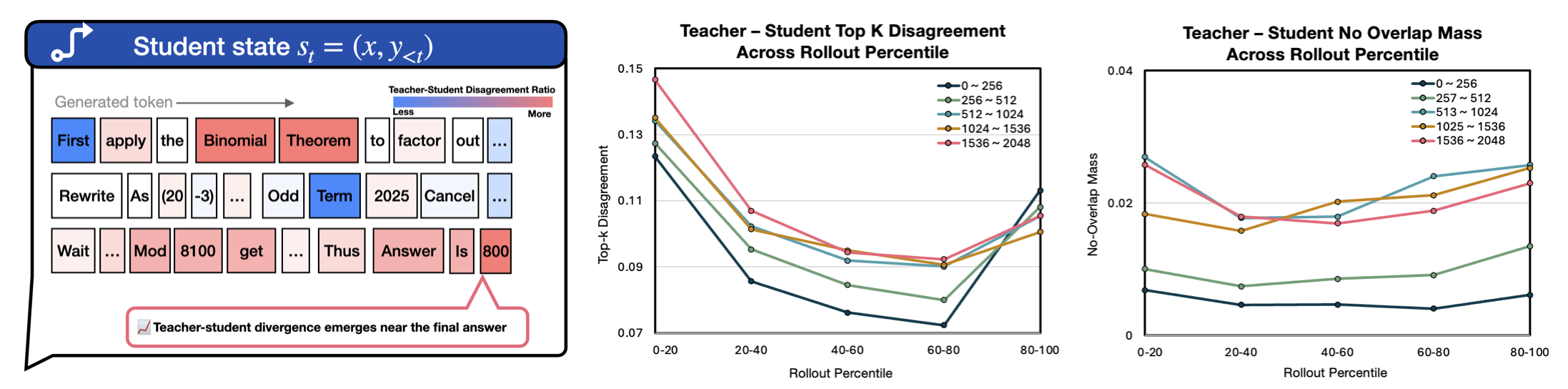}
\vspace{-0.55em}
\caption{\textbf{Target-reliability diagnostics.} \textbf{Left:} A token-level illustration of late-rollout teacher--student divergence, where disagreement concentrates near the final-answer region. \textbf{Middle:} Top-\(k\) disagreement rises again near the rollout tail and increases with privileged-context length. \textbf{Right:} No-overlap mass shows the same tail-end elevation, indicating that the teacher assigns probability mass to tokens outside the student's local predictive support.}
\label{fig:target_reliability_diagnostics}
\vspace{-0.45em}
\end{figure}

To quantify distributional misalignment, we measure Top-\(k\) Disagreement (\(D_k\)) and the Support Gap (\(M_{\tau}^{\mathrm{out}}\)). We compute these diagnostics on 1000 NuminaMath-style examples by sampling student rollouts from the public prompt and rescoring the same generated prefixes under the student, partial-view teacher, and full-view teacher prompts. We report top-\(k\) disagreement with \(k=16\), and compute support gap as the teacher mass assigned outside the student's locally plausible support; detailed estimator settings are provided in \Cref{app:diagnostic_estimator}. As shown in the middle and right panels of \Cref{fig:target_reliability_diagnostics}, both metrics experience a tail-end elevation and scale with privileged-context length; the left panel illustrates the same late-rollout divergence pattern at the token level. Notably, for sequences exceeding 1024 tokens, the support gap increases substantially in the final 40\% of the rollout. This tail-end elevation is consistent with \textbf{hindsight-induced manifold drift}. One explanation is that in early reasoning steps, the student and teacher distributions often share common ground, but as generation deepens, the student's prefix can accumulate slight deviations. The full-view teacher, conditioned on the gold future, remains anchored to the oracle trajectory. Rather than offering a localized correction, it assigns high probability to tokens that are justified by hindsight but logically disconnected from the student's actual accumulated prefix. Consequently, absolute full-view targets can pull the student toward \textbf{locally unsupported states}. Under the forward KL \(D_{\mathrm{KL}}(q \parallel \pi_\theta)=\sum_v q(v\mid s_t)\log\frac{q(v\mid s_t)}{\pi_\theta(v\mid s_t)}\), low-probability student terms \(\pi_\theta(v \mid s_t)\) in the denominator can disproportionately dominate the expected loss. This can bias optimization toward oracle shortcuts instead of facilitating constructive, step-by-step reasoning transfer, consistent with recent analyses showing that privileged teacher-only signals can induce information leakage and that successful OPD depends on locally shared high-probability tokens at student-visited states~\citep{selfdistilledrlvr2026,li2026rethinkingopd}.

\FloatBarrier
\begin{figure}[!htbp]
\centering
\includegraphics[width=0.88\textwidth]{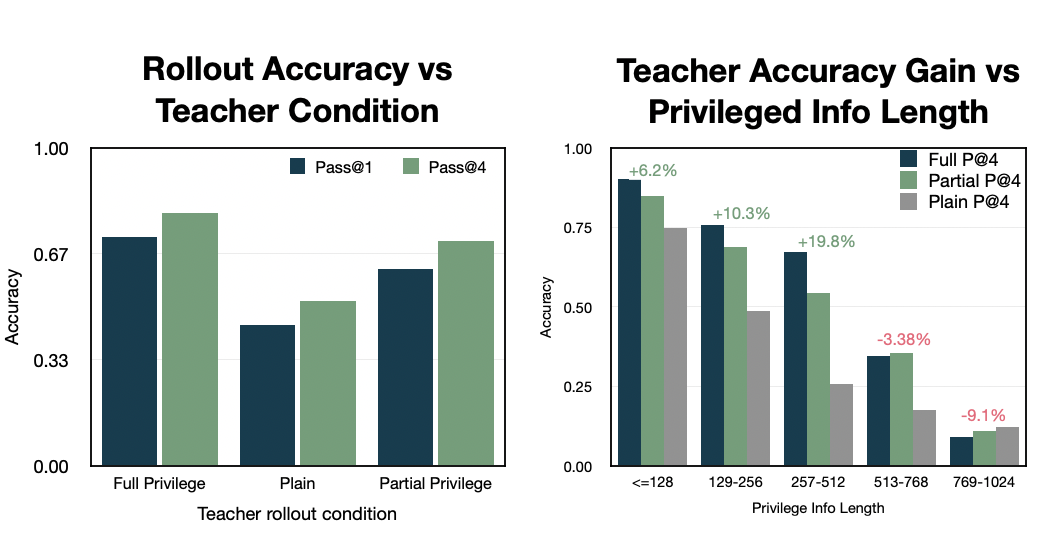}
\vspace{-0.4em}
\caption{\textbf{Reliability of privileged teachers in long-horizon contexts.} \textbf{Left:} Even with gold traces, the privileged teacher is imperfect and its reliability decays sharply as privileged context length increases. \textbf{Right:} The marginal gain of full-view over partial-view conditioning peaks at 256--512 tokens but then turns into negative marginal returns (up to $-9.1\%$) in longer horizons. This highlights that full-view targets in long-horizon reasoning are often noisier than partial-view anchors.}
\label{fig:teacher_reliability}
\vspace{-0.5em}
\end{figure}
\FloatBarrier

\subsection{Signal Fragility: The Illusion of Full-View Reliability}

Gold privileged context improves average teacher behavior, but it does not make the full-view teacher a uniformly reliable token target. On 5,000 NuminaMath problems with \llmname{Qwen2.5-7B-Instruct}~\citep{qwen25}, we run four rollouts for each teacher/student condition and extract final answers for scoring. \Cref{fig:teacher_reliability} shows that full-view teacher pass@4 drops from 90.18\% in the \(<128\)-token bucket to 9.09\% in the 768--1024-token bucket, and its marginal advantage over the partial view becomes negative by as much as \(-9.1\%\). This length sensitivity is consistent with recent OPD analyses showing that dense teacher targets can become locally incompatible when teacher-side information exceeds the student's reachable support~\citep{li2026rethinkingopd,fu2026revisitingopd,shenfeld2026sdft,penaloza2026privileged}.

The signal decomposition sharpens this point. Using the advantage scores defined in \Cref{eq:oracle_advantage}, \Cref{fig:signal_utility} shows that the partial privileged view is nearly as predictive of final correctness as the full-vs-student total advantage signal: partial advantage reaches ROC-AUC 0.766, close to the total signal at 0.779. By contrast, the marginal full-minus-partial signal falls to 0.585, indicating that the residual alone is a much weaker correctness classifier. Thus full-view information should not be treated as a standalone target; when used, it is better transferred as a controlled residual over a reliable partial anchor.

\section{AR-OPD: Anchored Residual Guidance}
\label{sec:method}

AR-OPD replaces full-view imitation with a dual-view target. The method evaluates two teachers on the same student-generated state \(s_t\): a partial teacher \(\qpart(\cdot\mid s_t)\) and a full teacher \(\qfull(\cdot\mid s_t)\). \Cref{fig:architecture} summarizes the target construction. The design follows a broader post-training pattern in which unstable long-horizon objects are decomposed into an anchor plus a controlled update, rather than optimized as a single monolithic target~\citep{zhu2025asft,li2025agentflow,yang2026gopd}.

\subsection{Partial Privilege Defines the Local Anchor}

Let the privileged information be \(z=(z_1,\ldots,z_L)\). We construct a partial privileged context by deterministic truncation:
\begin{equation}
z_{\mathrm{part}} = z_{1:\lfloor \rho L\rfloor},
\qquad
\rho\in(0,1).
\label{eq:partial_context}
\end{equation}
The default setting is \(\rho=0.5\). The partial teacher is
\begin{equation}
\qpart(\cdot\mid s_t)=\teacher(\cdot\mid x,z_{\mathrm{part}},y_{<t}).
\label{eq:qpart}
\end{equation}
The partial context exposes enough information to provide a structured setup while withholding later solution fragments, making the partial teacher a local anchor rather than an answer-leaking target. The front-partial choice is deliberately conservative: it resembles guided-prefix approaches that use early solution information to improve reachability, while avoiding direct supervision on the complete oracle trace~\citep{qu2026pope,shenfeld2026sdft,penaloza2026privileged}.

\subsection{Full Privilege Enters as a Scaled Residual}

The full teacher is
\begin{equation}
\qfull(\cdot\mid s_t)=\teacher(\cdot\mid x,z,y_{<t}).
\label{eq:qfull}
\end{equation}
In log-probability space, the full-view learning signal decomposes as
\begin{equation}
\log \qfull-\log\student
=
(\log \qpart-\log\student)
+
(\log \qfull-\log \qpart).
\label{eq:decomposition}
\end{equation}
The first term is the anchoring component, and the second term is the residual effect of giving the teacher the remaining privileged information.

\begin{aropdtakeawaybox}
\textbf{Anchored residual target.}
\begin{equation}
\boxed{
\begin{aligned}
\log \qlambda(v\mid s_t)
&=
\underbrace{\log \qpart(v\mid s_t)}_{\textcolor{aropdanchor}{\text{partial anchor}}}
+
\lambda
\underbrace{\big(\log \qfull(v\mid s_t)-\log \qpart(v\mid s_t)\big)}_{\textcolor{aropdresidual}{\text{full-minus-partial residual}}}
-
C_t .
\end{aligned}
}
\label{eq:anchored_residual}
\end{equation}
\end{aropdtakeawaybox}

\begin{aropdmethodbox}
\textbf{Target construction at each student state \(s_t\).}
\begin{enumerate}[leftmargin=1.8em,itemsep=-0.1em,topsep=0.2em]
\item Query the \anchorname, \(\qpart(\cdot\mid s_t)\), using the front partial privileged trace.
\item Query the \fullname, \(\qfull(\cdot\mid s_t)\), using the complete privileged trace.
\item Transfer only \(\lambda(\log\qfull-\log\qpart)\), then normalize to obtain \(\qlambda\).
\item Distill the student against \(\qlambda\) on the same on-policy prefix \(s_t\).
\end{enumerate}
\end{aropdmethodbox}

The normalizer \(C_t\) makes \(\qlambda(\cdot\mid s_t)\) a valid distribution. The form makes the object-level change explicit: imitation no longer targets \(\qfull\) directly; it targets an anchored belief update around \(\qpart\). The student is trained by KL distillation to the anchored residual target:
\begin{equation}
\mathcal{L}_{\mathrm{AR-OPD}}(\theta)
=
\mathbb{E}_{x\sim\mathcal{D},\,y\sim\student}
\left[
\sum_{t=1}^{T}
\kl\big(\qlambda(\cdot\mid s_t)\parallel\student(\cdot\mid s_t)\big)
\right].
\label{eq:AR-OPD_loss}
\end{equation}
In implementation, targets are built on student-generated rollouts rather than fixed teacher traces. For each sampled prefix, the partial and full teacher prompts rescore the same next-token state with temperature \(1.0\); the anchored residual distribution is then formed in log space and optimized with forward KL. We use the same prompt and completion limits as the evaluation setup, and apply a small reference mixup and residual clipping only for numerical stability; the full configuration is listed in \Cref{tab:training_config_provenance}.

\begin{figure}[!t]
\centering
\vspace{-0.85em}
\includegraphics[width=0.72\linewidth,trim=0 10 0 0,clip]{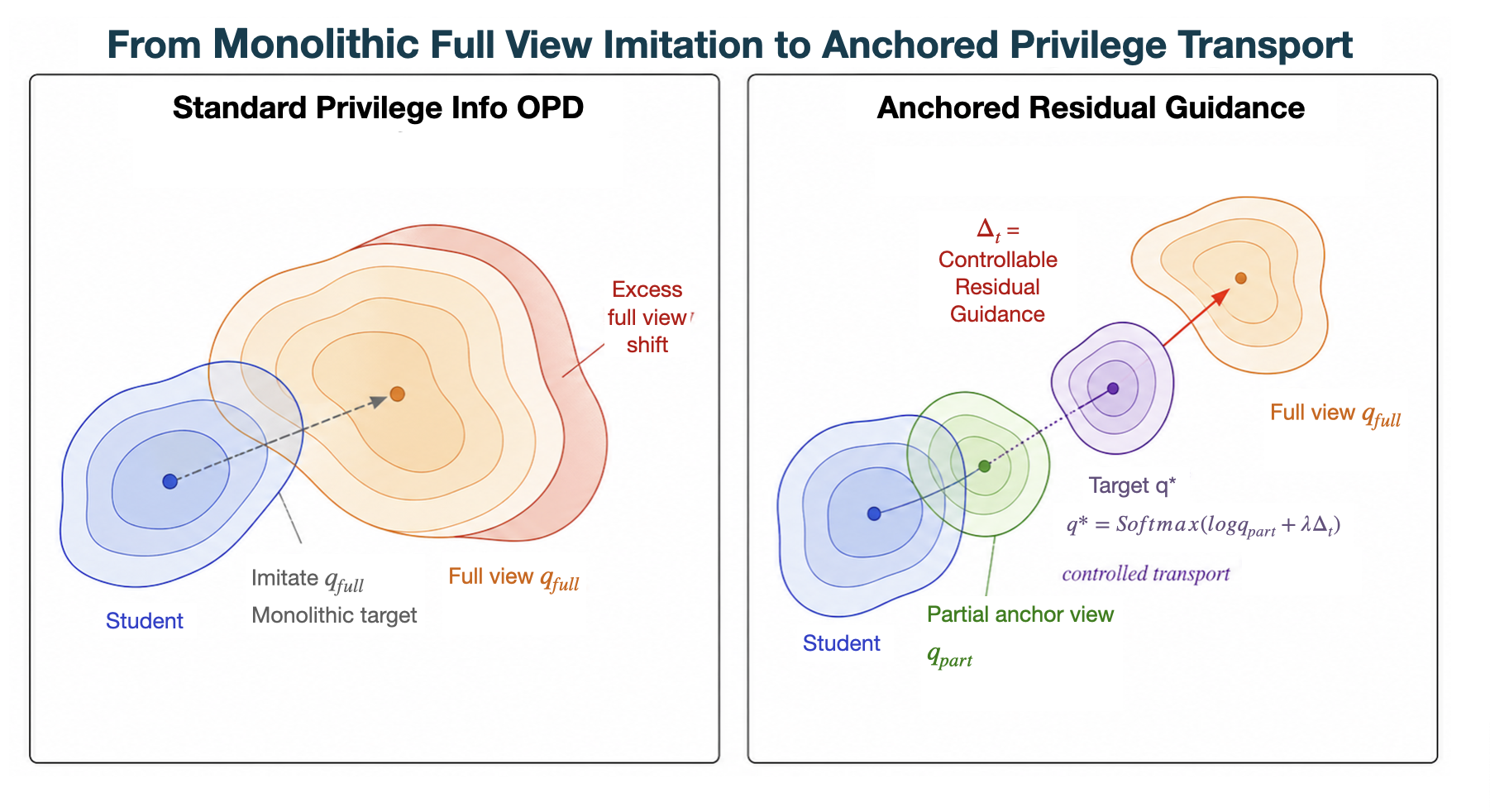}
\vspace{-0.95em}
\caption{\textbf{Constructing reachable targets via anchored residual guidance.} AR-OPD anchors on the student-compatible partial view and uses the full view only as a controlled directional residual.}
\label{fig:distribution_shift}
\vspace{0.55em}
\end{figure}

The coefficient \(\lambda\) controls the amount of full-view residual transfer. When \(\lambda=0\), AR-OPD becomes partial privileged OPD; when \(\lambda=1\), it recovers the full privileged target by construction. Thus \(\lambda<1\) is a contractive residual regime that keeps the target anchored between the partial and full views, while \(\lambda>1\) extrapolates beyond the full-view log shift. \Cref{fig:distribution_shift} visualizes this change: the full view is used as a directional residual around the partial anchor rather than as an absolute imitation target. The formulation makes \(\lambda\) closer to a residual-guidance strength than to a generic smoothing parameter, complementing work that studies reward extrapolation, target reformulation, and entropy-aware OPD control~\citep{yang2026gopd,jang2026veto,jin2026eaopd}.

\FloatBarrier
\vspace{0.25em}

\section{Experiments: Performance Gains Follow Target Reliability}
\label{sec:experiments}

The evaluation separates outcome evidence from mechanism evidence through four checks: cross-domain outcome, anchor compatibility, residual value, and training dynamics.

\subsection{Experimental Setup}

\paragraph{Tasks and datasets.} We evaluate four reasoning task groups. For mathematics, we train on NuminaMath-style CoT data~\citep{numinamath2024} and evaluate on MATH500~\citep{math,lightman2023let} and AMC23~\citep{maa2023amc}. For code, we train on Magicoder-Evol-Instruct data~\citep{wei2024magicoder,luo2023wizardcoder} and evaluate HumanEval and MBPP with pass@1~\citep{chen2021evaluating,austin2021program}. For science, we use SciKnowEval Chemistry L-3~\citep{feng2024sciknoweval} with exact-match multiple-choice accuracy. For medical QA, we train on MedMCQA~\citep{pal2022medmcqa} and evaluate on MedQA~\citep{jin2020medqa} and MedCQA~\citep{pal2022medmcqa}. We use the language-model evaluation harness for standardized evaluation plumbing where applicable~\citep{eval-harness}.

\paragraph{Model and training.} Unless stated otherwise, we use \llmname{Qwen2.5-7B-Instruct} as the base model~\citep{qwen25}. The student and teacher views share the same model family and differ only in the privileged context exposed to the teacher during target construction. Across tasks, we set the maximum completion length to 1024 tokens and use a 2048-token context window including the input prompt. The default partial rate is \(\rho=0.5\). We sweep \(\lambda\in\{0.4,0.6,0.8,1.0,1.2\}\) to study residual strength, keeping other hyperparameters fixed across methods.

\paragraph{Compared methods.} We compare the base model, supervised fine-tuning (SFT), full privileged OPD, partial privileged OPD, and AR-OPD. The comparison separates demonstration-only adaptation, full privileged self-distillation, and guided partial-privilege training~\citep{ouyang2022instructgpt,shenfeld2026sdft,qu2026pope}. Full OPD trains directly against \(\qfull\) with an EMA self-teacher. AR-OPD trains against \(\qlambda\), the anchor-residual target.

\paragraph{Evaluation protocol.} All reported methods are evaluated with the same public prompts and without privileged context at inference time. Privileged information is used only to construct training targets, so the comparison isolates how different target constructions affect the same base model. For math and multiple-choice tasks, we extract the final answer with the task-specific evaluator; for code, we use the standard unit-test pass@1 protocol.
All distillation variants use the same on-policy rollout budget, optimizer settings, prompt limits, and final-checkpoint reporting rule. Thus, differences in the main table reflect how the target distribution is constructed rather than changes in data scale, inference-time context, or evaluator choice.

\subsection{Main Results: Performance}

\begin{table*}[!t]
\centering
\caption{\textbf{Anchored residual guidance achieves the strongest average performance across domains.} Metrics are accuracy or pass@1; code, science, and medical results remain part of the main cross-domain evidence, while checkpointed math-only diagnostics are separated in \Cref{app:math}.}
\label{tab:main_results}
\renewcommand{\arraystretch}{1.12}
\setlength{\tabcolsep}{3.5pt}
\small
\begin{tabular}{ll>{\columncolor{aropdtablebase}}c>{\columncolor{aropdtablesft}}c c>{\columncolor{aropdtablefull}}c>{\columncolor{aropdtableours}}c}
\toprule
\textbf{Domain} & \textbf{Benchmark} & \textbf{Base} & \textbf{SFT} & \textbf{Partial OPD} & \fullterm{\textbf{Full OPD}} & \anchorterm{\textbf{AR-OPD}} \\
\midrule
\multirow{2}{*}{Math} & MATH500 & 61.8 & 63.2 & 70.6 & 71.0 & \hlfirst{\textbf{74.6}} \\
& AMC23 & 45.0 & 47.5 & 45.0 & 55.0 & \hlfirst{\textbf{57.5}} \\
\midrule
\multirow{2}{*}{Code} & HumanEval & 79.2 & 76.8 & 79.4 & 78.8 & \hlfirst{\textbf{82.6}} \\
& MBPP & 77.5 & 79.4 & 77.8 & 78.5 & \hlfirst{\textbf{80.4}} \\
\midrule
Science & SciKnowEval & 32.12 & 47.4 & 64.2 & 68.0 & \hlfirst{\textbf{68.5}} \\
\midrule
\multirow{2}{*}{Medical} & MedQA & 58.7 & 66.0 & 64.3 & 65.2 & \hlfirst{\textbf{66.7}} \\
& MedCQA & 53.2 & 56.7 & 55.4 & 59.0 & \hlfirst{\textbf{61.5}} \\
\midrule
\rowcolor{aropdtableavg}
\textbf{Average} & All benchmarks & 58.2 & 62.4 & 65.2 & 67.9 & \hlfirst{\textbf{70.3}} \\
\bottomrule
\end{tabular}
\end{table*}
\FloatBarrier

\paragraph{Performance.}
AR-OPD consistently improves over standard SFT and competitive on-policy distillation baselines across reasoning-intensive and knowledge-heavy domains. As shown in \Cref{tab:main_results}, AR-OPD obtains the best average score of 70.3, improving over the base model by \textbf{12.1 points}, SFT by \textbf{7.9 points}, and the strong Full OPD baseline by \textbf{2.3 points}. The gains are largest in mathematical reasoning, where hindsight-conditioned targets are most likely to induce support mismatch: AR-OPD reaches 74.6 on MATH500 and 57.5 on AMC23, exceeding Full OPD by \textbf{3.6 and 2.5 points}, respectively. Notably, even Partial OPD, which remains blinded to the final answer, outperforms standard full-trace SFT by an average of 2.8 points. While Partial OPD lacks destination awareness, AR-OPD adds residual guidance over a locally valid partial anchor, yielding a \textbf{5.1-point average gain over Partial OPD}. The improvement extends beyond math: AR-OPD also ranks first on HumanEval, MBPP, SciKnowEval, MedQA, and MedCQA.

\begin{figure}[!htbp]
\centering
\vspace{-0.25em}
\includegraphics[width=\linewidth,trim=4 10 4 0,clip]{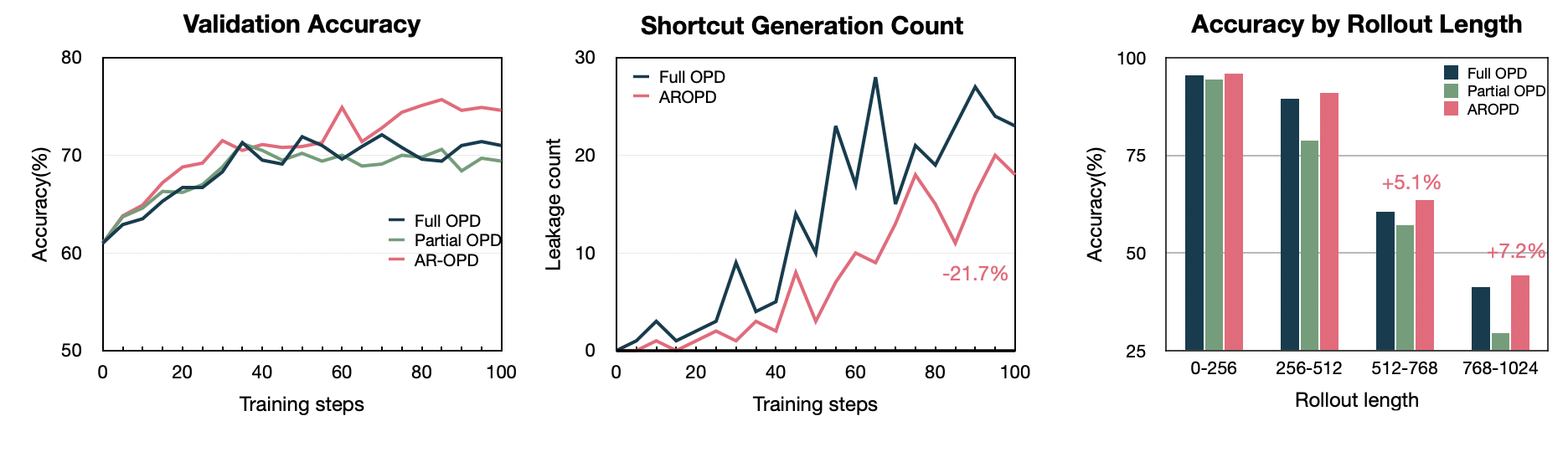}
\vspace{-0.35em}
\caption{\textbf{AR-OPD improves validation accuracy, reduces shortcuts, and is strongest on long rollouts.} \textbf{Left:} Validation accuracy during Math10K training: AR-OPD shows consistent learning gains, while Full OPD peaks early and then largely plateaus. \textbf{Middle:} Shortcut-generation count over training steps; at the final training step (100\%), AR-OPD has 18 shortcut events versus 23 for Full OPD, a 21.7\% relative reduction. \textbf{Right:} Accuracy grouped by rollout length, where AR-OPD improves over Full OPD by 5.1 points on 512--768-token rollouts and 7.2 points on 768--1024-token rollouts.}
\label{fig:shortcut_accuracy}
\vspace{-0.5em}
\end{figure}
\FloatBarrier

\subsection{Shortcut Dynamics and Long-Rollout Accuracy}
\label{sec:shortcut_dynamics}

\paragraph{Shortcut dynamics.}
The shortcut diagnostic quantifies the failure mode from \Cref{fig:motivation}. By anchoring to the partial view, AR-OPD consistently suppresses shortcut events, reducing the average count across all 21 training checkpoints from 12.95 for Full OPD to 7.52. At the final checkpoint, AR-OPD produces only 18 shortcuts compared to Full OPD's 23, a \textbf{21.7\% relative reduction}.

\paragraph{Long-horizon accuracy.}
AR-OPD's advantage scales with trajectory length (\Cref{fig:shortcut_accuracy}, right). While competitive on short rollouts, it surpasses Full OPD by \textbf{5.1 points} on 512--768-token rollouts and \textbf{7.2 points} on 768--1024-token rollouts. These results suggest that, as reasoning horizons grow, anchored residual targets reduce the compounding hindsight drift that can degrade monolithic full-view imitation.

\AROPDhighlight{The results show that controlling the full-view residual improves over full privileged OPD, while the gain over Partial OPD confirms that the residual itself adds useful destination-directed signal.}

\subsection{Lambda Sweep and Residual Control}
\label{sec:lambda_sweep}

\begin{table}[!h]
\centering
\caption{\textbf{Lambda sweep for the AR-OPD residual scale on the Math-10K checkpointed run.} Metrics are relaxed accuracy. Contractive residual transfer (\(\lambda<1\)) gives the strongest settings, while extrapolating past the full-view residual with \(\lambda=1.2\) degrades performance.}
\label{tab:lambda_sweep}
\small
\setlength{\tabcolsep}{5pt}
\begin{tabular}{lccc}
\toprule
\textbf{Residual scale} & \textbf{Regime} & \textbf{Math500} & \textbf{AMC23} \\
\midrule
\rowcolor{aropdtableours}
\(\lambda=0.4\) & Weak residual & 71.2 & 50.0 \\
\rowcolor{aropdtableours}
\(\lambda=0.6\) & Contractive residual & 71.6 & \hlfirst{\textbf{57.5}} \\
\rowcolor{aropdtableours}
\(\lambda=0.8\) & Contractive residual & \hlfirst{\textbf{74.6}} & 55.0 \\
\rowcolor{aropdtablefull}
\(\lambda=1.0\) & Full-view limit & 71.8 & 55.0 \\
\rowcolor{aropdtableavg}
\(\lambda=1.2\) & Extrapolated residual & 69.0 & 50.0 \\
\bottomrule
\end{tabular}
\end{table}

Residual scale determines how aggressively AR-OPD transfers the full-view log shift. By \Cref{eq:anchored_residual}, \(\lambda=1\) recovers \(\qfull\), values below one keep the target contractively tied to the partial anchor, and values above one extrapolate beyond the full-view direction. The sweep in \Cref{tab:lambda_sweep} shows that \(\lambda=0.8\) is strongest on Math500 and \(\lambda=0.6\) is strongest on AMC23, whereas the weak residual setting \(\lambda=0.4\) underuses answer-directed guidance and the extrapolated setting \(\lambda=1.2\) underperforms both contractive settings. We therefore treat \(\lambda\) as the main control knob balancing partial-anchor stability against full-view over-transfer, in line with recent OPD and RL-for-reasoning work showing that target scale and reformulation materially affect optimization behavior~\citep{yang2026gopd,jin2026eaopd,wang2025beyond}.

\section{Discussion: Decomposing Privileged Distribution Shift}
\label{sec:discussion}
Our results frame privilege as a target-design problem rather than a binary choice between using or discarding oracle information. Full-view oracle distributions can provide useful destination-directed signal, but absolute imitation can create a hindsight trap: tokens valid under the privileged future may remain unsupported from the student's current prefix. The diagnostics suggest that this mismatch grows near rollout tails and that longer privileged context does not necessarily improve teacher reliability. AR-OPD mitigates this by separating supervision into a locally valid anchor and a controlled residual, turning full-view information into a bounded update rather than a standalone coordinate.

\paragraph{Future work.}
Promising directions include adaptive anchors based on student uncertainty or gradient variance, broader privileged modalities beyond gold traces such as search heuristics or human preferences, and long-horizon settings such as multi-turn agents or repository-level coding where hindsight interference may be more pronounced.

\section{Limitations}
\label{sec:limitations}

AR-OPD requires two teacher forward passes per student state, increasing target-generation cost relative to single-view OPD and motivating comparison with search- or verifier-based alternatives~\citep{generalist_rm,deepcritic,walder2025pass}. Our experiments use one open model family, \llmname{Qwen2.5-7B-Instruct}, and modest data sizes; broader sweeps over data, scale, and families such as Llama~\citep{touvron2023llama} remain important. Algorithmically, we use a fixed partial anchor (\(\rho=0.5\)) and static residual scale; adaptive schedules, diagnostics beyond NuminaMath, and structured-output robustness remain open~\citep{evalplus,dont-overthink,madaan2023self,yuan2024self}.

\clearpage
\bibliographystyle{plainnat}
\bibliography{dopd_refs}

@article{hinton2015distilling,
  title={Distilling the Knowledge in a Neural Network},
  author={Hinton, Geoffrey and Vinyals, Oriol and Dean, Jeff},
  journal={arXiv preprint arXiv:1503.02531},
  year={2015}
}

@inproceedings{ross2011dagger,
  title={A Reduction of Imitation Learning and Structured Prediction to No-Regret Online Learning},
  author={Ross, Stephane and Gordon, Geoffrey J. and Bagnell, J. Andrew},
  booktitle={International Conference on Artificial Intelligence and Statistics},
  year={2011}
}

@article{ouyang2022instructgpt,
  title={Training Language Models to Follow Instructions with Human Feedback},
  author={Ouyang, Long and Wu, Jeffrey and Jiang, Xu and Almeida, Diogo and Wainwright, Carroll L. and Mishkin, Pamela and Zhang, Chong and Agarwal, Sandhini and Slama, Katarina and Ray, Alex and others},
  journal={Advances in Neural Information Processing Systems},
  year={2022}
}

@article{schulman2017ppo,
  title={Proximal Policy Optimization Algorithms},
  author={Schulman, John and Wolski, Filip and Dhariwal, Prafulla and Radford, Alec and Klimov, Oleg},
  journal={arXiv preprint arXiv:1707.06347},
  year={2017}
}

@misc{taori2023alpaca,
  title={Stanford Alpaca: An Instruction-following LLaMA Model},
  author={Taori, Rohan and Gulrajani, Ishaan and Zhang, Tianyi and Dubois, Yann and Li, Xuechen and Guestrin, Carlos and Liang, Percy and Hashimoto, Tatsunori B.},
  year={2023},
  howpublished={\url{https://crfm.stanford.edu/2023/03/13/alpaca.html}}
}

@article{touvron2023llama,
  title={LLaMA: Open and Efficient Foundation Language Models},
  author={Touvron, Hugo and Lavril, Thibaut and Izacard, Gautier and Martinet, Xavier and Lachaux, Marie-Anne and Lacroix, Timothee and Roziere, Baptiste and Goyal, Naman and Hambro, Eric and Azhar, Faisal and Rodriguez, Aurelien and Joulin, Armand and Grave, Edouard and Lample, Guillaume},
  journal={arXiv preprint arXiv:2302.13971},
  year={2023}
}

@article{guo2025deepseekr1,
  title={DeepSeek-R1: Incentivizing Reasoning Capability in LLMs via Reinforcement Learning},
  author={Guo, Daya and Yang, Dejian and Zhang, Haowei and Song, Junxiao and Zhang, Ruoyu and Xu, Runxin and Zhu, Qihao and Ma, Shirong and Wang, Peiyi and Bi, Xiao and others},
  journal={arXiv preprint arXiv:2501.12948},
  year={2025}
}

@misc{thinkingmachines2025opd,
  title={On-Policy Distillation},
  author={{Thinking Machines Lab}},
  year={2025},
  howpublished={\url{https://thinkingmachines.ai/blog/on-policy-distillation/}}
}

@article{gu2023minillm,
  title={MiniLLM: Knowledge Distillation of Large Language Models},
  author={Gu, Yuxian and Dong, Li and Wei, Furu and Huang, Minlie},
  journal={arXiv preprint arXiv:2306.08543},
  year={2023}
}

@article{zhao2026opsd,
  title={Self-Distilled Reasoner: On-Policy Self-Distillation for Large Language Models},
  author={Zhao, Siyan and Xie, Zhihui and Liu, Mengchen and Huang, Jing and Pang, Guan and Chen, Feiyu and Grover, Aditya},
  journal={arXiv preprint arXiv:2601.18734},
  year={2026}
}

@article{fu2026revisitingopd,
  title={Revisiting On-Policy Distillation: Empirical Failure Modes and Simple Fixes},
  author={Fu, Yuqian and Huang, Haohuan and Jiang, Kaiwen and Liu, Jiacai and Jiang, Zhuo and Zhu, Yuanheng and Zhao, Dongbin},
  journal={arXiv preprint arXiv:2603.25562},
  year={2026}
}

@article{li2026rethinkingopd,
  title={Rethinking On-Policy Distillation of Large Language Models: Phenomenology, Mechanism, and Recipe},
  author={Li, Yaxuan and Zuo, Yuxin and He, Bingxiang and Zhang, Jinqian and Xiao, Chaojun and Qian, Cheng and Yu, Tianyu and Gao, Huan-ang and Yang, Wenkai and Liu, Zhiyuan and Ding, Ning},
  journal={arXiv preprint arXiv:2604.13016},
  year={2026}
}

@article{qu2026pope,
  title={POPE: Learning to Reason on Hard Problems via Privileged On-Policy Exploration},
  author={Qu, Yuxiao and Setlur, Amrith and Smith, Virginia and Salakhutdinov, Ruslan and Kumar, Aviral},
  journal={arXiv preprint arXiv:2601.18779},
  year={2026}
}

@article{yang2026gopd,
  title={Learning beyond Teacher: Generalized On-Policy Distillation with Reward Extrapolation},
  author={Yang, Wenkai and Liu, Weijie and Xie, Ruobing and Yang, Kai and Yang, Saiyong and Lin, Yankai},
  journal={arXiv preprint arXiv:2602.12125},
  year={2026}
}

@article{li2025agentflow,
  title={In-the-Flow Agentic System Optimization for Effective Planning and Tool Use},
  author={Li, Zhuofeng and Zhang, Haoxiang and Han, Seungju and Liu, Sheng and Xie, Jianwen and Zhang, Yu and Choi, Yejin and Zou, James and Lu, Pan},
  journal={arXiv preprint arXiv:2510.05592},
  year={2025}
}

@article{shenfeld2026sdft,
  title={Self-Distillation Enables Continual Learning},
  author={Shenfeld, Idan and Damani, Mehul and H{\"u}botter, Jonas and Agrawal, Pulkit},
  journal={arXiv preprint arXiv:2601.19897},
  year={2026}
}

@article{zhu2025asft,
  title={Anchored Supervised Fine-Tuning},
  author={Zhu, He and Su, Junyou and Lai, Peng and Ma, Ren and Zhang, Wenjia and Yang, Linyi and Chen, Guanhua},
  journal={arXiv preprint arXiv:2509.23753},
  year={2025}
}

@article{jang2026veto,
  title={Stable On-Policy Distillation through Adaptive Target Reformulation},
  author={Jang, Ijun and Yeom, Jewon and Yeo, Juan and Lim, Hyunggu and Kim, Taesup},
  journal={arXiv preprint arXiv:2601.07155},
  year={2026}
}

@article{jin2026eaopd,
  title={Entropy-Aware On-Policy Distillation of Language Models},
  author={Jin, Woogyeol and Min, Taywon and Yang, Yongjin and Kadhe, Swanand Ravindra and Zhou, Yi and Wei, Dennis and Baracaldo, Nathalie and Lee, Kimin},
  journal={arXiv preprint arXiv:2603.07079},
  year={2026}
}

@article{penaloza2026privileged,
  title={Privileged Information Distillation for Language Models},
  author={Penaloza, Emiliano and Vattikonda, Dheeraj and Gontier, Nicolas and Lacoste, Alexandre and Charlin, Laurent and Caccia, Massimo},
  journal={arXiv preprint arXiv:2602.04942},
  year={2026}
}

@article{selfdistilledrlvr2026,
  title={Self-Distilled RLVR},
  author={Yang, Chenxu and Qin, Chuanyu and Si, Qingyi and Chen, Minghui and Gu, Naibin and Yao, Dingyu and Lin, Zheng and Wang, Weiping and Wang, Jiaqi and Duan, Nan},
  journal={arXiv preprint arXiv:2604.03128},
  year={2026}
}

@misc{numinamath2024,
  title={NuminaMath},
  author={{AI-MO/Numina}},
  year={2024},
  howpublished={\url{https://huggingface.co/datasets/AI-MO/NuminaMath-CoT}}
}

@misc{maa2023amc,
  title={2023 American Mathematics Competitions Problems},
  author={{Mathematical Association of America}},
  year={2023},
  howpublished={\url{https://maa.org/math-competitions/american-mathematics-competitions-amc/}}
}

@article{jin2020medqa,
  title={What Disease does this Patient Have? A Large-scale Open Domain Question Answering Dataset from Medical Exams},
  author={Jin, Di and Pan, Eileen and Oufattole, Nassim and Weng, Wei-Hung and Fang, Hanyi and Szolovits, Peter},
  journal={arXiv preprint arXiv:2009.13081},
  year={2020}
}

@article{austin2021program,
  title={Program Synthesis with Large Language Models},
  author={Austin, Jacob and Odena, Augustus and Nye, Maxwell and Bosma, Maarten and Michalewski, Henryk and Dohan, David and Jiang, Ellen and Cai, Carrie and Terry, Michael and Le, Quoc and Sutton, Charles},
  journal={arXiv preprint arXiv:2108.07732},
  year={2021}
}

@inproceedings{wei2024magicoder,
  title={Magicoder: Empowering Code Generation with {OSS}-Instruct},
  author={Wei, Yuxiang and Wang, Zhe and Liu, Jiawei and Ding, Yifeng and Zhang, Lingming},
  booktitle={Proceedings of the 41st International Conference on Machine Learning},
  pages={52632--52657},
  year={2024},
  volume={235},
  series={Proceedings of Machine Learning Research},
  month={21--27 Jul},
  publisher={PMLR},
  url={https://proceedings.mlr.press/v235/wei24h.html}
}

@misc{luo2023wizardcoder,
  title={WizardCoder: Empowering Code Large Language Models with Evol-Instruct},
  author={Luo, Ziyang and Xu, Can and Zhao, Pu and Sun, Qingfeng and Geng, Xiubo and Hu, Wenxiang and Tao, Chongyang and Ma, Jing and Lin, Qingwei and Jiang, Daxin},
  year={2023},
  eprint={2306.08568},
  archivePrefix={arXiv},
  primaryClass={cs.CL},
  url={https://arxiv.org/abs/2306.08568}
}

@inproceedings{pal2022medmcqa,
  title={MedMCQA: A Large-scale Multi-Subject Multi-Choice Dataset for Medical Domain Question Answering},
  author={Pal, Ankit and Umapathi, Logesh Kumar and Sankarasubbu, Malaikannan},
  booktitle={Proceedings of the Conference on Health, Inference, and Learning},
  pages={248--260},
  year={2022},
  volume={174},
  series={Proceedings of Machine Learning Research},
  publisher={PMLR},
  url={https://proceedings.mlr.press/v174/pal22a.html}
}

@inproceedings{agarwal2024policy,
  title={On-policy distillation of language models: Learning from self-generated mistakes},
  author={Agarwal, Rishabh and Vieillard, Nino and Zhou, Yongchao and Stanczyk, Piotr and Garea, Sabela Ramos and Geist, Matthieu and Bachem, Olivier},
  booktitle={The Twelfth International Conference on Learning Representations},
  year={2024}
}

@article{lamb2016professor,
  title={Professor forcing: A new algorithm for training recurrent networks},
  author={Lamb, Alex M and ALIAS PARTH GOYAL, Anirudh Goyal and Zhang, Ying and Zhang, Saizheng and Courville, Aaron C and Bengio, Yoshua},
  journal={Advances in neural information processing systems},
  volume={29},
  year={2016}
}

@article{brown2020language,
  title={Language models are few-shot learners},
  author={Brown, Tom and Mann, Benjamin and Ryder, Nick and Subbiah, Melanie and Kaplan, Jared D and Dhariwal, Prafulla and Neelakantan, Arvind and Shyam, Pranav and Sastry, Girish and Askell, Amanda and others},
  journal={Advances in neural information processing systems},
  volume={33},
  pages={1877--1901},
  year={2020}
}

@article{wei2022chain,
  title={Chain-of-thought prompting elicits reasoning in large language models},
  author={Wei, Jason and Wang, Xuezhi and Schuurmans, Dale and Bosma, Maarten and Xia, Fei and Chi, Ed and Le, Quoc V and Zhou, Denny and others},
  journal={Advances in Neural Information Processing Systems},
  volume={35},
  pages={24824--24837},
  year={2022}
}

@article{lightman2023let,
  title={Let's Verify Step by Step},
  author={Lightman, Hunter and Kosaraju, Vineet and Burda, Yura and Edwards, Harri and Baker, Bowen and Lee, Teddy and Leike, Jan and Schulman, John and Sutskever, Ilya and Cobbe, Karl},
  journal={arXiv preprint arXiv:2305.20050},
  year={2023}
}

@article{madaan2023self,
  title={Self-refine: Iterative refinement with self-feedback},
  author={Madaan, Aman and Tandon, Niket and Gupta, Prakhar and Hallinan, Skyler and Gao, Luyu and Wiegreffe, Sarah and Alon, Uri and Dziri, Nouha and Prabhumoye, Shrimai and Yang, Yiming and others},
  journal={arXiv preprint arXiv:2303.17651},
  year={2023}
}

@article{yuan2024self,
  title={Self-rewarding language models},
  author={Yuan, Weizhe and Pang, Richard Yuanzhe and Cho, Kyunghyun and Sukhbaatar, Sainbayar and Xu, Jing and Weston, Jason},
  journal={arXiv preprint arXiv:2401.10020},
  year={2024}
}

@article{feng2024sciknoweval,
  title={Sciknoweval: Evaluating multi-level scientific knowledge of large language models},
  author={Feng, Kehua and Ding, Keyan and Wang, Weijie and Zhuang, Xiang and Wang, Zeyuan and Qin, Ming and Zhao, Yu and Yao, Jianhua and Zhang, Qiang and Chen, Huajun},
  journal={arXiv preprint arXiv:2406.09098},
  year={2024}
}

@article{chen2021evaluating,
  title={Evaluating large language models trained on code},
  author={Chen, Mark and Tworek, Jerry and Jun, Heewoo and Yuan, Qiming and Pinto, Henrique Ponde De Oliveira and Kaplan, Jared and Edwards, Harri and Burda, Yuri and Joseph, Nicholas and Brockman, Greg and others},
  journal={arXiv preprint arXiv:2107.03374},
  year={2021}
}

@misc{eval-harness,
  author       = {Gao, Leo and Tow, Jonathan and Abbasi, Baber and Biderman, Stella and Black, Sid and DiPofi, Anthony and Foster, Charles and Golding, Laurence and Hsu, Jeffrey and Le Noac'h, Alain and Li, Haonan and McDonell, Kyle and Muennighoff, Niklas and Ociepa, Chris and Phang, Jason and Reynolds, Laria and Schoelkopf, Hailey and Skowron, Aviya and Sutawika, Lintang and Tang, Eric and Thite, Anish and Wang, Ben and Wang, Kevin and Zou, Andy},
  title        = {The Language Model Evaluation Harness},
  month        = 07,
  year         = 2024,
  publisher    = {Zenodo},
  version      = {v0.4.3},
  doi          = {10.5281/zenodo.12608602},
  url          = {https://zenodo.org/records/12608602}
}

@article{walder2025pass,
  title={Pass@ K Policy Optimization: Solving Harder Reinforcement Learning Problems},
  author={Walder, Christian and Karkhanis, Deep},
  journal={arXiv preprint arXiv:2505.15201},
  year={2025}
}

@article{wang2025beyond,
  title={Beyond the 80/20 Rule: High-Entropy Minority Tokens Drive Effective Reinforcement Learning for LLM Reasoning},
  author={Wang, Shenzhi and Yu, Le and Gao, Chang and Zheng, Chujie and Liu, Shixuan and Lu, Rui and Dang, Kai and Chen, Xionghui and Yang, Jianxin and Zhang, Zhenru and others},
  journal={arXiv preprint arXiv:2506.01939},
  year={2025}
}

@article{deepcritic,
  title={Deepcritic: Deliberate critique with large language models},
  author={Yang, Wenkai and Chen, Jingwen and Lin, Yankai and Wen, Ji-Rong},
  journal={arXiv preprint arXiv:2505.00662},
  year={2025}
}

@article{rlhf-workflow,
  title={RLHF Workflow: From Reward Modeling to Online RLHF},
  author={Dong, Hanze and Xiong, Wei and Pang, Bo and Wang, Haoxiang and Zhao, Han and Zhou, Yingbo and Jiang, Nan and Sahoo, Doyen and Xiong, Caiming and Zhang, Tong},
  journal={arXiv preprint arXiv:2405.07863},
  year={2024}
}

@misc{yang2025qwen3,
  title={Qwen3 Technical Report},
  author={{Qwen Team}},
  year={2025},
  eprint={2505.09388},
  archivePrefix={arXiv},
  primaryClass={cs.CL},
  url={https://arxiv.org/abs/2505.09388}
}

@misc{xiao2026mimo,
  title={MiMo: Unlocking the Reasoning Potential of Language Model -- From Pretraining to Posttraining},
  author={{Xiaomi MiMo Team}},
  year={2025},
  eprint={2505.07608},
  archivePrefix={arXiv},
  primaryClass={cs.CL},
  url={https://arxiv.org/abs/2505.07608}
}

@misc{zeng2026glm5,
  title={GLM-5: from Vibe Coding to Agentic Engineering},
  author={{GLM-5 Team}},
  year={2026},
  eprint={2602.15763},
  archivePrefix={arXiv},
  primaryClass={cs.LG},
  url={https://arxiv.org/abs/2602.15763}
}

@inproceedings{ho2022classifier,
  title={Classifier-free diffusion guidance},
  author={Ho, Jonathan and Salimans, Tim},
  booktitle={NeurIPS 2022 Workshop on Score-Based Methods},
  year={2022}
}

@misc{bradley2024classifier,
  title={Classifier-Free Guidance is a Predictor-Corrector},
  author={Bradley, Arwen and Nakkiran, Preetum},
  year={2024},
  eprint={2408.09000},
  archivePrefix={arXiv},
  primaryClass={cs.LG},
  url={https://arxiv.org/abs/2408.09000}
}

@article{verl,
  title   = {HybridFlow: A Flexible and Efficient RLHF Framework},
  author  = {Guangming Sheng and Chi Zhang and Zilingfeng Ye and Xibin Wu and Wang Zhang and Ru Zhang and Yanghua Peng and Haibin Lin and Chuan Wu},
  year    = {2024},
  journal = {arXiv preprint arXiv: 2409.19256}
}

@article{generalist_rm,
  title={Inference-time scaling for generalist reward modeling},
  author={Liu, Zijun and Wang, Peiyi and Xu, Runxin and Ma, Shirong and Ruan, Chong and Li, Peng and Liu, Yang and Wu, Yu},
  journal={arXiv preprint arXiv:2504.02495},
  year={2025}
}

@inproceedings{seq-kd,
  title={Sequence-level knowledge distillation},
  author={Kim, Yoon and Rush, Alexander M},
  booktitle={Proceedings of the 2016 conference on empirical methods in natural language processing},
  pages={1317--1327},
  year={2016}
}

@article{distilbert,
  title={DistilBERT, a distilled version of BERT: smaller, faster, cheaper and lighter},
  author={Sanh, Victor and Debut, Lysandre and Chaumond, Julien and Wolf, Thomas},
  journal={arXiv preprint arXiv:1910.01108},
  year={2019}
}

@inproceedings{evalplus,
  title = {Is Your Code Generated by Chat{GPT} Really Correct? Rigorous Evaluation of Large Language Models for Code Generation},
  author = {Liu, Jiawei and Xia, Chunqiu Steven and Wang, Yuyao and Zhang, Lingming},
  booktitle = {Thirty-seventh Conference on Neural Information Processing Systems},
  year = {2023},
  url = {https://openreview.net/forum?id=1qvx610Cu7},
}

@article{qwen25,
  title={Qwen2.5 Technical Report},
  author={{Qwen Team}},
  journal={arXiv preprint arXiv:2412.15115},
  year={2024}
}

@article{dont-overthink,
  title={Don't Overthink it. Preferring Shorter Thinking Chains for Improved LLM Reasoning},
  author={Hassid, Michael and Synnaeve, Gabriel and Adi, Yossi and Schwartz, Roy},
  journal={arXiv preprint arXiv:2505.17813},
  year={2025}
}

@inproceedings{math,
  title={Measuring Mathematical Problem Solving With the {MATH} Dataset},
  author={Hendrycks, Dan and Burns, Collin and Kadavath, Saurav and Arora, Akul and Basart, Steven and Tang, Eric and Song, Dawn and Steinhardt, Jacob},
  booktitle={Thirty-fifth Conference on Neural Information Processing Systems Datasets and Benchmarks Track},
  year={2021},
  url={https://openreview.net/forum?id=7Bywt2mQsCe}
}

\FloatBarrier
\appendix
\section*{Use of LLMs}

Large language models were utilized as auxiliary tools to assist with code writing, manuscript editing, and figure preparation. The authors maintained full review and control over all research ideas, experimental designs, analyses, and final content. LLM assistance was used for LaTeX cleanup, wording refinement, caption compression, and drafting small analysis scripts during revision. LLMs were not used to generate benchmark measurements; all reported numbers come from the described experiments and were checked by the authors.

\section*{Code and Data Availability}

The project page is available at \url{https://vanhowe.github.io/AR-OPD/}. It provides access to the paper assets, training and evaluation code, configuration files, launch scripts, lightweight evaluation data, and data-layout manifests. Large derived training datasets, checkpoints, and generated report bundles are not bundled in the release; the released scripts document the expected local layout and validation path.

\setcounter{figure}{0}
\renewcommand{\thefigure}{A.\arabic{figure}}
\setcounter{table}{0}
\renewcommand{\thetable}{A.\arabic{table}}

\section{Prompt for Privileged Information}
\label{app:privileged_prompt}

For each training example, we construct a standard student prompt and two privileged teacher prompts. The student prompt contains only the original programming problem. The privileged prompts additionally include an example response before asking the model to produce its own response.

\begin{aropdpromptbox}
\textbf{Template for full and partial privileged information.}

\vspace{0.35em}
{\small\ttfamily
[Original problem]\par
\vspace{0.45em}
This is an example for a response to the question:\par
[Privileged response]\par
\vspace{0.45em}
Now answer with a response of your own, including the thinking process.\par
}
\end{aropdpromptbox}

For the full privileged condition, \texttt{[Privileged response]} is the complete reference trace, including the final code answer when available. For the partial privileged condition, \texttt{[Privileged response]} is a partial trace that removes the final answer. In the code data, the partial trace is constructed by removing fenced code blocks and inline code snippets from the reference response, then truncating the remaining explanation to the first 50\% of characters without splitting a word.

\begin{aropdpromptbox}
\textbf{Resulting training record.}
\begin{itemize}[leftmargin=1.4em,itemsep=0.15em,topsep=0.35em]
\item \texttt{prompt}: the original programming problem;
\item \texttt{partial\_teacher\_prompt}: the original problem plus partial privileged information;
\item \texttt{full\_teacher\_prompt}: the original problem plus full privileged information;
\item \texttt{gold\_trace}: the full reference response;
\item \texttt{partial\_trace}: the partial privileged response;
\item \texttt{gold\_answer}: the extracted final code answer.
\end{itemize}
\end{aropdpromptbox}

\section{Training and Evaluation Provenance}
\label{app:provenance}

To make the main comparison reproducible, \Cref{tab:training_data_provenance,tab:training_config_provenance,tab:evaluation_provenance} summarize the training data, shared optimization configuration, and evaluation protocol used for \Cref{tab:main_results}. Unless otherwise specified, all methods report the final checkpoint under the same evaluation protocol.

\begin{table*}[!t]
\centering
\caption{\textbf{Training data and privileged-view construction.} Each row summarizes the source and dual-view construction used to build the corresponding training records.}
\label{tab:training_data_provenance}
\small
\setlength{\tabcolsep}{3pt}
\begin{tabularx}{\textwidth}{L{0.14\textwidth} C{0.08\textwidth} L{0.27\textwidth} X}
\toprule
\textbf{Domain} & \textbf{Train size} & \textbf{Source} & \textbf{View construction} \\
\midrule
Math & 10K & NuminaMath-style CoT subset & Partial teacher keeps the first 50\% of the reasoning trace at a complete-word boundary and removes the final answer; full teacher keeps the complete trace with the gold final answer. \\
\midrule
Code & 4K & Magicoder-Evol-Instruct-110K~\citep{wei2024magicoder}; upstream Evol-CodeAlpaca-style data & Partial trace removes fenced code blocks and inline code snippets, then keeps the first 50\% of the remaining explanation without splitting a word; full trace keeps the complete reference response. \\
\midrule
Science & 2K & SciKnowEval Chemistry L-3~\citep{feng2024sciknoweval} & Multiple-choice demonstrations use the same dual-view construction; accuracy is computed by exact match on the final answer choice. \\
\midrule
Medical & 2K & MedMCQA subset~\citep{pal2022medmcqa} & Multiple-choice demonstrations use the same dual-view construction; accuracy is computed by exact match on the final answer choice. \\
\bottomrule
\end{tabularx}
\end{table*}

\begin{table}[!t]
\centering
\caption{\textbf{Shared dual-GPU serving and training configuration.} Code uses 4K training examples; science and medical use 2K examples.}
\label{tab:training_config_provenance}
\small
\setlength{\tabcolsep}{4pt}
\begin{tabular}{ll}
\toprule
\textbf{Parameter} & \textbf{Value} \\
\midrule
Base model & \llmname{Qwen2.5-7B-Instruct} \\
Hardware topology & GPU 0 training, GPU 1 vLLM teacher serving \\
Learning rate / scheduler & \(2\times10^{-5}\), cosine, warmup ratio 0.1 \\
Epochs / max steps & 1 epoch, epoch-based training \\
Per-device batch / global batch & 1 / 64 \\
Gradient accumulation & 64 \\
Prompt / completion length & max prompt 1024, max completion 1024 \\
Context window & vLLM max model length 2048, including input prompt \\
Training temperature & 1.0 \\
Precision & bf16 when GPU is available \\
Distillation objective & forward KL, distill alpha 0.0 \\
Ref mixup / residual clip & mixup alpha 0.01, residual clip 5.0 \\
Default residual scale & \(\lambda=0.6\), with sweep in \Cref{tab:lambda_sweep} \\
Teacher CPU offload & enabled via dual-teacher CPU offload \\
\bottomrule
\end{tabular}
\end{table}

\begin{table*}[!t]
\centering
\caption{\textbf{Evaluation protocol for the main results.} All methods report final-checkpoint performance under the same evaluator and prompt protocol.}
\label{tab:evaluation_provenance}
\small
\setlength{\tabcolsep}{4pt}
\begin{tabularx}{\textwidth}{L{0.18\textwidth} L{0.18\textwidth} L{0.24\textwidth} X}
\toprule
\textbf{Benchmark} & \textbf{Decoding} & \textbf{Evaluator} & \textbf{Metric} \\
\midrule
MATH500 & Greedy, temperature 0.0 & Math answer parser with relaxed normalization & Relaxed accuracy \\
AMC23 & Greedy, temperature 0.0 & Math answer parser with relaxed normalization & Relaxed accuracy \\
HumanEval & Greedy, temperature 0.0 & Original HumanEval evaluator~\citep{chen2021evaluating} & pass@1 \\
MBPP & Greedy, temperature 0.0 & Original MBPP evaluator~\citep{austin2021program} & pass@1 \\
SciKnowEval & Greedy, temperature 0.0 & Exact match on final multiple-choice answer & Accuracy \\
MedQA & Greedy, temperature 0.0 & Exact match on final multiple-choice answer & Accuracy \\
MedCQA & Greedy, temperature 0.0 & Exact match on final multiple-choice answer & Accuracy \\
\bottomrule
\end{tabularx}
\end{table*}

\section{Diagnostic Definitions}
\label{app:diagnostic_definitions}

\subsection{Connection to Classifier-Free Guidance}
\label{app:cfg_connection}

Classifier-Free Guidance (CFG) combines unconditional and conditional score estimates by scaling the residual effect of a condition~\citep{ho2022classifier}. In diffusion models, this can be written abstractly as
\begin{equation}
\tilde{s}_{\gamma}(x_t,c)
=
s_{\theta}(x_t,\varnothing)
+
\gamma\big(
s_{\theta}(x_t,c)-s_{\theta}(x_t,\varnothing)
\big),
\label{eq:cfg_logic}
\end{equation}
where the unconditional score provides a reference trajectory and the conditional residual steers sampling. Recent analyses further connect CFG to predictor-corrector behavior~\citep{bradley2024classifier}. We use this only as a loose analogy: AR-OPD does not import a diffusion objective, but it adopts the same structural separation between an anchor distribution and a scaled directional residual.

\subsection{Diagnostic Estimator}
\label{app:diagnostic_estimator}

We estimate the target-reliability diagnostics on 1000 NuminaMath-style math examples. For each example, we sample a student rollout from the public prompt and treat every generated prefix as a student-visited state \(s_t\). We then rescore the same generated token sequence under the public student prompt, partial privileged teacher prompt, and full privileged teacher prompt using the same checkpoint. Logits are converted to probabilities with temperature \(T=1.0\); generation uses top-\(p=1.0\), maximum generation length 1024, and maximum prompt length 2048. We report top-\(k\) disagreement with \(k=16\), averaged over token positions and grouped by five rollout-percentile bins and privileged-context-length buckets.

For a teacher distribution \(q\), we first define the top-\(k\) overlap ratio as
\begin{equation}
O_k(q,\student;s_t)
=
\frac{
\left|\operatorname{Top}_k(q(\cdot\mid s_t))\cap
\operatorname{Top}_k(\student(\cdot\mid s_t))\right|
}{k}.
\label{eq:topk_overlap}
\end{equation}
The corresponding top-\(k\) disagreement ratio is
\begin{equation}
D_k(q,\student;s_t)
=
1-O_k(q,\student;s_t).
\label{eq:topk_disagreement}
\end{equation}
A larger \(D_k\) indicates weaker local alignment between the teacher's high-probability tokens and the student's reachable predictive support on the same prefix.

The implementation-level diagnostic also records shared teacher mass over the overlapping top-\(k\) set:
\begin{equation}
\widehat{B}_k(q,\student;s_t)
=
\sum_{v\in
\operatorname{Top}_k(q(\cdot\mid s_t))
\cap
\operatorname{Top}_k(\student(\cdot\mid s_t))}
q(v\mid s_t).
\label{eq:topk_shared_mass}
\end{equation}
The corresponding top-\(k\) no-overlap mass proxy is the teacher mass assigned to top-\(k\) teacher tokens that do not appear in the student's top-\(k\) set:
\begin{equation}
\widehat{N}_k(q,\student;s_t)
=
\sum_{v\in
\operatorname{Top}_k(q(\cdot\mid s_t))
\setminus
\operatorname{Top}_k(\student(\cdot\mid s_t))}
q(v\mid s_t).
\label{eq:topk_no_overlap_proxy}
\end{equation}
This top-\(k\) proxy should be distinguished from the full-vocabulary thresholded support gap below.

We also define the teacher mass outside the student's local predictive support at threshold \(\tau\):
\begin{equation}
M_{\tau}^{\mathrm{out}}(q,\student;s_t)
=
\sum_{v:\student(v\mid s_t)<\tau} q(v\mid s_t).
\label{eq:support_gap}
\end{equation}
This no-overlap mass, also referred to as the support gap, measures how much probability the teacher assigns to tokens that the student currently treats as locally implausible. Equivalently, since \(q\) is normalized, \(M_{\tau}^{\mathrm{out}}=1-\sum_{v:\student(v\mid s_t)\ge\tau}q(v\mid s_t)\). Thus the plotted no-overlap quantity is already the complement of in-support teacher mass. High values indicate that the teacher distribution may be correct under its privileged view but locally difficult for the student to imitate from \(s_t\).

\subsection{Teacher-Signal Advantage Scores}
\label{app:signal_advantage}

To measure which part of the privileged signal predicts final-answer correctness, we compute log-probability advantage scores over a token span \(\mathcal{T}\). The total, partial, and marginal advantages are
\begin{align}
A_{\mathrm{full}}(\mathcal{T})
&=
\sum_{t\in\mathcal{T}}
\left(
\log \qfull(y_t\mid s_t)-\log \student(y_t\mid s_t)
\right),
\notag\\
A_{\mathrm{part}}(\mathcal{T})
&=
\sum_{t\in\mathcal{T}}
\left(
\log \qpart(y_t\mid s_t)-\log \student(y_t\mid s_t)
\right),
\notag\\
A_{\mathrm{marg}}(\mathcal{T})
&=
A_{\mathrm{full}}(\mathcal{T})-A_{\mathrm{part}}(\mathcal{T})
=
\sum_{t\in\mathcal{T}}
\left(
\log \qfull(y_t\mid s_t)-\log \qpart(y_t\mid s_t)
\right).
\label{eq:oracle_advantage}
\end{align}
We use each scalar score as a predictor of final-answer correctness and report ROC-AUC in \Cref{fig:signal_utility}.

\subsection{Shortcut-Event Counting Protocol}
\label{app:shortcut_detection}

To quantify privileged-information leakage in \Cref{fig:shortcut_accuracy}, we evaluate trained privileged-OPD checkpoints on a held-out NuminaMath validation split and generate 5K responses per checkpoint. We use a deterministic response-level detector for shortcut-like evidence, such as references to unavailable solutions, answer-conditioned certainty without local derivation, or prompt-external answer cues. A response is counted once if any rule fires, so the reported count is a conservative response-level diagnostic of target-side leakage rather than a judge of mathematical correctness.

\section{Math Deep Dive and Training Diagnostics}
\label{app:math}

This appendix reports supplemental diagnostics for the math setting; these results do not replace the main cross-domain comparison in \Cref{tab:main_results}. The Numina500 split exposes checkpoint and residual-scale sensitivity rather than redefining the headline benchmark table.

\subsection{Teacher-Signal Diagnostics}

\begin{wrapfigure}[22]{r}{0.50\textwidth}
\vspace{-0.6em}
\centering
\includegraphics[width=0.96\linewidth]{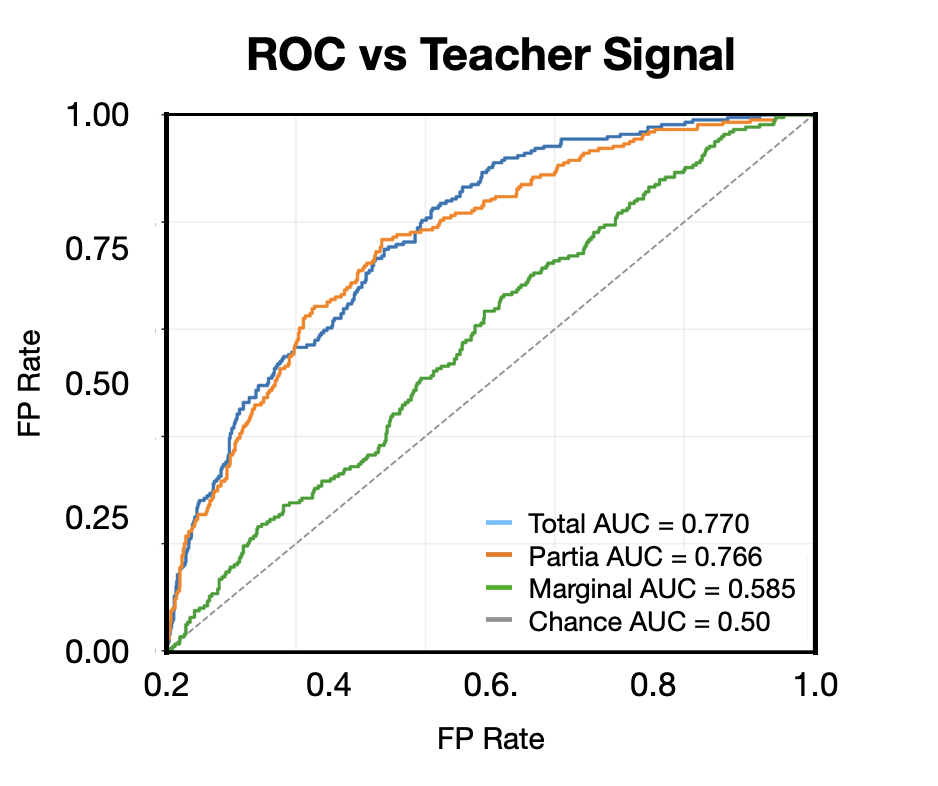}
\vspace{-0.25em}
\caption{\textbf{Correctness-predictive teacher signal.} The partial privileged signal nearly matches the total signal in ROC-AUC, while the marginal full-minus-partial residual is substantially weaker.}
\label{fig:signal_utility}
\vspace{-0.2em}
\end{wrapfigure}

\Cref{fig:signal_utility} and \Cref{fig:full_vs_partial_support} provide the appendix-level diagnostics behind the main mechanism claim: the partial view carries stable correctness-predictive signal, while the full-view target creates larger disagreement and support gaps against the student.

\FloatBarrier

\subsection{Checkpoint-Level Math Results}

The checkpoint sweep is used only as supplemental evidence for residual-scale sensitivity: contractive AR-OPD settings are strongest on the headline math diagnostics, while the main cross-domain comparison remains the decisive result.

\subsection{Failure Modes}

Some raw failures reflect length and answer-format instability; relaxed-grading gains, especially for the partial-only model, should be read as formatting sensitivity rather than reasoning improvement.

\clearpage
\begin{center}
\vspace*{-1.2em}
\makebox[\linewidth][c]{\includegraphics[width=1.08\linewidth]{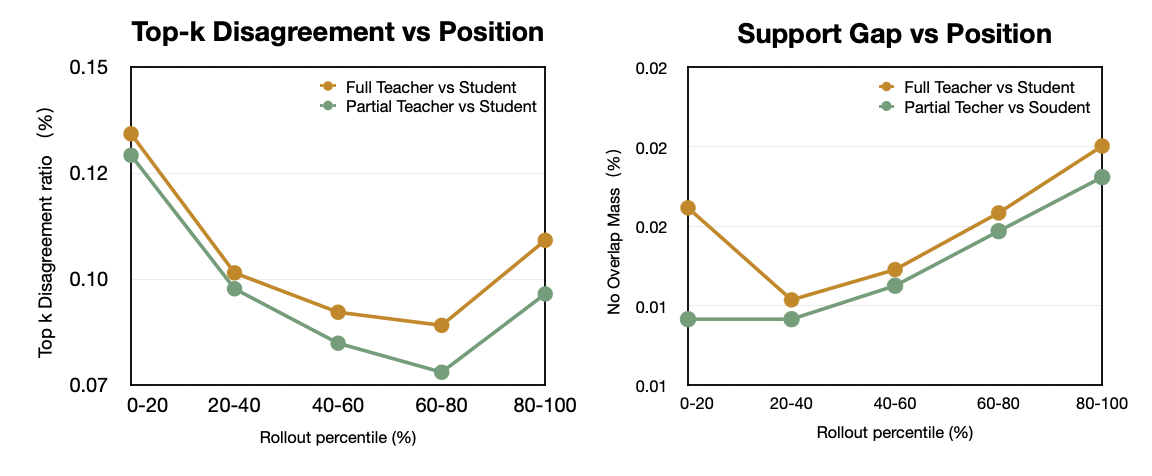}}
\vspace{-0.6em}
\captionof{figure}{\textbf{Teacher--student support-gap diagnostics.} Full-view targets show higher top-\(k\) disagreement and no-overlap mass than partial-view anchors across rollout positions.}
\label{fig:full_vs_partial_support}
\end{center}

\end{document}